# Optimal Motion Generation of the Bipedal Under-Actuated Planar Robot for Stair Climbing

Aref Amiri [1], Hassan Salarieh [2]

[1]Graduate Student, Sharif University of Technology, Tehran; aref.amiri@mech.sharif.edu
[2]Professor, Sharif University of Technology, Tehran; salarieh@sharif.edu

**Abstract**
The importance of humanoid robots in today's world is undeniable, one of the most important features of humanoid robots is the ability to maneuver in environments such as stairs that other robots can not easily cross. A suitable algorithm to generate the path for the bipedal robot to climb is very important. In this paper, an optimization-based method to generate an optimal stairway for under-actuated bipedal robots without an ankle actuator is presented. The generated paths are based on zero and non-zero dynamics of the problem, and according to the satisfaction of the zero dynamics constraint in the problem, tracking the path is possible, in other words, the problem can be dynamically feasible. The optimization method used in the problem is a gradient-based method that has a suitable number of function evaluations for computational processing. This method can also be utilized to go down the stairs.

**Introduction**
Inspired by human body physics, bipedal robots have many degrees of freedom and can perform various actions with their joint movements. Bipedal robots can adapt to different environments that other wheeled robots are unable to move. The study of path (trajectory) generation methods as a reference for the output of the control problem of bipedal robots in this regard is essential. For the bipedal robot to climb the stairs, it is necessary to analyze the movement of them ascending the stairs and to examine the method of planning the bipedal robot to move and to determine the position of feet for walking on the stairs [1].

So far, researches have been done on how to go up and downstairs and find a suitable or optimal path for bipedal robots. Various papers using optimization algorithms and considering the robot angles as polynomial functions tried to design an optimal path for a 6-degree bipedal robot [2]. Some articles have even paths planned for multi-legged robots to cross the stairs [3]. Some articles also used stability criteria such as ZMP in designing their paths [4-7]. But this method is only appliable for robots that have feet (soles) with ankle joint actuators, which often have much lower speed in maneuvering than under-actuated robots without feet, and of course, due to the relatively large feet have more wasted energy. Some articles also derive their initial path using data based on motion capturing and then try to optimize their results by combining optimization methods [8]. However, according to the existing literature, few articles have attempted to design a holonomic path for under-actuated bipedal robots without feet. Due to the importance of optimal motion planning, a lot of work has been done in recent years in this area.

In this paper, the problem of motion planning is investigated to find the optimal paths for under-actuated bipedal robots to step on the stairs, the results obtained as a control output will cause the robot to move properly and optimally. This article consists of three sections. In the first part, the dynamic model of the bipedal robot is derived. In the second part, the constraints of the optimization problem are examined, in the third part, the cost function and method of optimal problem solving and finding a suitable movement gate are examined. In the fourth section, the results are presented and discussed, and at the end, the research of this article is summarized as the conclusion.

**Dynamics equation**
The dynamic model of the robot is shown in Figure 1. The robot has 7 degrees of freedom and 5 links, each leg has two joints (one in the knee and the other in the hip) and 3 degrees of freedom. We assume that the contact of the tip of the leg is the point.

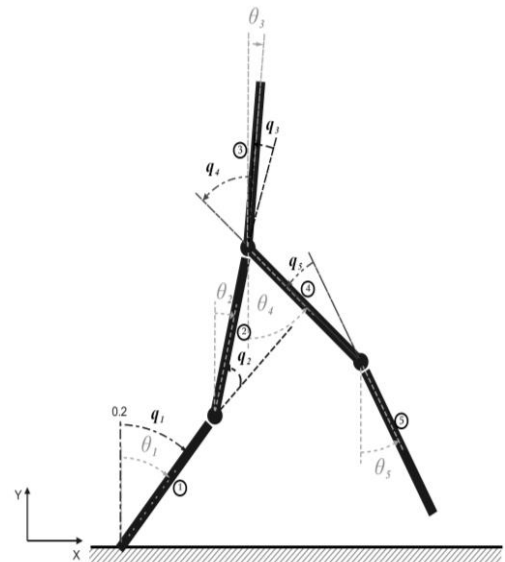

Figure 1. Planar bipedal robot

The robot's motion is planar and the robot has 4 actuators, two actuators at the knees and two actuators at the junction of the hip and the trunk so that there is one actuator between each leg and trunk. It is assumed that by hitting the tip of the swing leg on the ground, the other leg rises from the ground, in other words, the robot has no double support phase. So, when moving on the stairs, no time is wasted for placing both feet on the ground. Therefore, the hybrid dynamic equations of a robot are a combination of a single support phase and collision phase. The equations of the hybrid model are as follows:

$$\Sigma : \begin{cases} \dot{x} = f(x) + g(x)u & x^- \notin \Gamma \\ x^+ = \Delta(x^-) & x^- \in \Gamma \end{cases} \quad (1)$$

The vector $x := (q^T, \dot{q}^T)^T$ consists of the vector of generalized coordinates and their derivatives. $\Delta$ is a map to find the states of the system exactly after the collision, and the positive and negative symbols indicate the states of the system before and after the collision. The switch condition is as follows:

$$\Gamma = \left\{ (q, \dot{q}) \in x \mid P_2^v(q) = 0, P_2^h(q) > 0 \right\} \quad (2)$$

In equation (2), $P_2^h$ represents the horizontal position of the swing leg and $P_2^v$ represents its vertical position.

The dynamic equations of the robot before and after the collision and in the single support phase can be written as follows:

$$M_{(q)}\ddot{q} + C_{(q,\dot{q})}\dot{q} + G_{(q)} = B_{(q)}u \quad (3)$$

Matrix B is also a pre-multiplication matrix in the torque vector and is not a square matrix due to the under-actuation of the system.

In Equation 1, there is an expression called zero dynamics, and it is easy to separate this term if the generalized coordinates of the system are written in relative terms (as has been done in this paper). The satisfaction of this constraint is important in two ways. First, if this constraint is not satisfied, the problem of optimizing the input torques is practically ambiguous, because these torques are not really applicable to the problem. Although it may lead to a feasible kinematic equation (kinematically possible), it is not feasible in terms of control (open-loop), i.e. it is not dynamically possible.

**Optimization problem**
The most important constraint of the problem, called zero dynamics, was introduced in the previous section. Other constraints in this issue are important to plan the robot movement in the best way; the constraints of the optimization problem are generally classified into two general modes of constraints based on dynamics and constraints based on kinematics.

1. Dynamic constraints:
Torque limit: because the torque generators have a certain limit (inequality constraint).
Zero dynamic: the importance of which was mentioned earlier (equality constraint).

Coefficient of friction limit: for the robot to move on real environments, the ratio of horizontal force to vertical force should not be more or less than a certain limit. In other words, the coefficient of friction required for stepping should not exceed a certain limit that can not be implemented in real environments. (inequality constraint).

2. Kinematic constraints:
Configuration: As an initial and final condition, the robot needs to move from an initial configuration to a final configuration. The best option is for the initial and final state to be the same so that the robot has periodicity in its movement and the best footprint is in the middle of each stair Figure 2 (equality constraint).

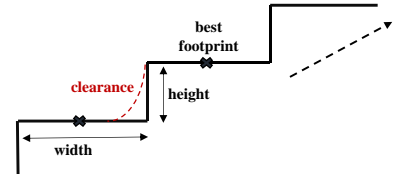

Figure 2. Stair properties

Angular velocity limit: Because motors have limited angular velocity production. (inequality constraint)
Contact in single support phase: The robot is in contact with the ground during the single support phase and the acceleration of the contact point in the horizontal and vertical direction during this period is zero. (equality constraint)
Swing leg collision: The robot swing leg during the single-phase phase, except at the beginning and end of the phase, should not collide with the ground, on the other hand, should have a suitable distance to the obstacles.
Knees movement limitation: To create maximum similarity to human movement, the robot knee should not be opened and closed too much.
Failure to satisfy any of the above constraints will cause problems in creating optimal and appropriate movement.

**Optimization method**
This optimization is a nonlinear, constrained, and single-objective problem.
Cost function: To find the optimal path, various cost functions are considered, for example, the norm of torque input, system input energy, and cost of transport are common options. In this paper, we consider the norm of torque inputs as the cost function. By this choice, the torques are rational in size and will have proper distribution (If the optimization problem is solved properly).

$$J = \int_0^T (\sum_{i=0}^{4} u_i^2(\tau)) \, d\tau \quad (4)$$

In the above equation, $T$ is the length of the time period.

Selection of optimization variables: Optimization variables can have different types, one of the best choices is the paths followed by generalized coordinates. Here our choice is a time-varying path as a function of polynomials. The polynomial functions are uniform and smooth, and they are also simple for deriving.

$$q_k(t) = \sum_{i=0}^{n=4} \alpha_{k,i} t^i \quad (5)$$

The degree of this polynomial must be chosen in such a way that the number of optimization parameters, which are the same as the number of polynomial coefficients, are appropriate (minimum value to have a smooth motion satisfied the mentioned constraints). In this article, we choose the function of order 4 to have freedom of action in terms of the optimization problem and also not to make the number of optimization parameters of the problem irrational and complicated.

Method of solving the optimization problem: This optimization problem is solved by Variable Metric methods for constrained optimization. This method is a gradient-based method, which provides a desirable and fast solution. Another advantage of this method is to not get out easily from the feasible area [9].

**Results and Discussion**
Following the model and algorithm presented above, a bipedal robot has been simulated to climb the stairs. The height of the stairs is considered 20cm and the width of the stairs is 40cm. The robot model specifications are in accordance with Table 1. The initial and final angles of the bipedal robot as a configuration are given in Table 2. Here the initial and final configurations are intuitively obtained from the human configuration. The speed of crossing each step is .5 seconds. The torque limit applied to the system is 150 N.m and the maximum angular velocity of the motors 10 rad/sec can be.

Table 1. Rabbit robot properties [10]

| Symbol | Value |
|---|---|
| $m_1, m_5$ | 3.2 kg |
| $m_2, m_4$ | 6.8 kg |
| $m_3$ | 20 kg |
| $I_1, I_5$ | 0.93 kg-m$^2$ |
| $I_2, I_4$ | 1.08 kg-m$^2$ |
| $I_3$ | 2.22 kg-m$^2$ |
| $l_1, l_5$ | 0.4 m |
| $l_2, l_4$ | 0.4 m |
| $l_3$ | 0.625 m |
| $d_1, d_5$ | 0.128 m |
| $d_2, d_4$ | 0.163 m |
| $d_3$ | 0.2 m |

Table 2. The initial and final configuration

| Parameters | Initial value(rad) | Final value(rad) |
|---|---|---|
| $q_1$ | 0.2618 | 0.1964 |
| $q_2$ | 1.3140 | 0 |
| $q_3$ | -1.2267 | 0.0219 |
| $q_4$ | -0.0219 | 1.2267 |
| $q_5$ | 0 | 1.3140 |

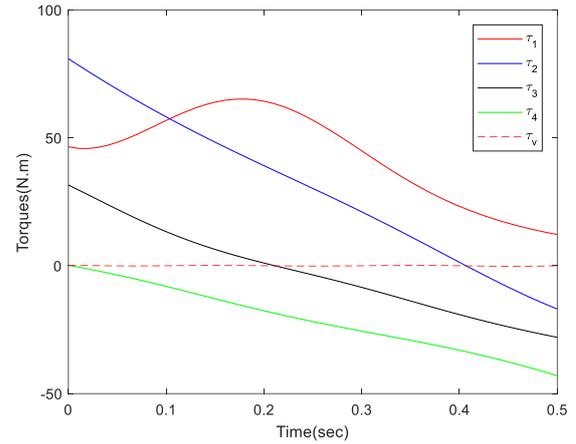

Figure 3. Input torques

According to Figure 3, the torques have a good margin from the saturation and compared to other articles and research reviewed in the introduction, more optimal results have been obtained, also zero dynamics ($\tau_v$) in a very good way is satisfied.

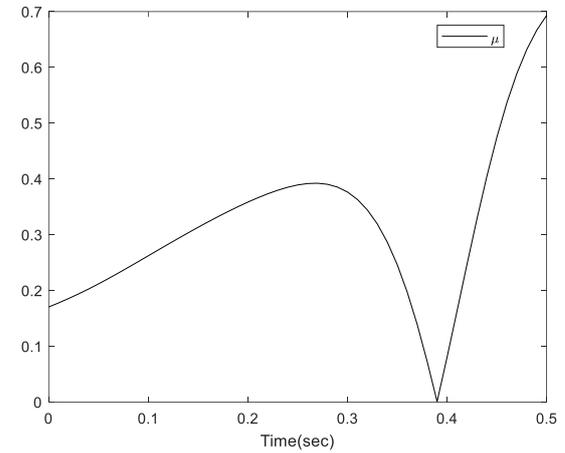

Figure 4. Friction coefficient

According to Figure 4, it is clear that the generated path needs the maximum coefficient of friction .69 to slip, so on all surfaces that have a coefficient of friction higher than .69 there is the ability to move.

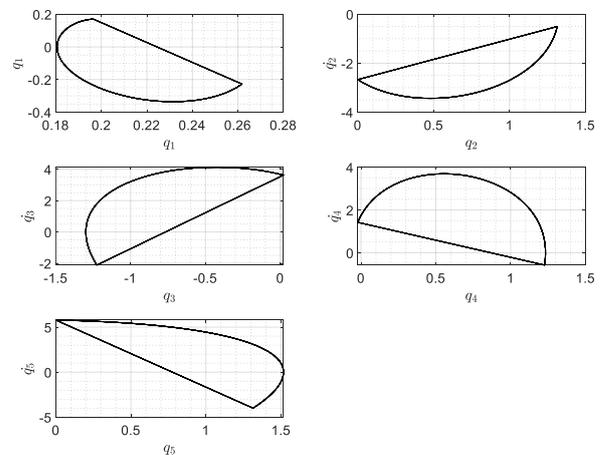

Figure 5. Angles vs. angular velocities

According to Figure 5, the generated paths, due to the nature of the polynomial functions, have a smooth and non-breaking behavior, and the angular velocities are far from their saturation limit.

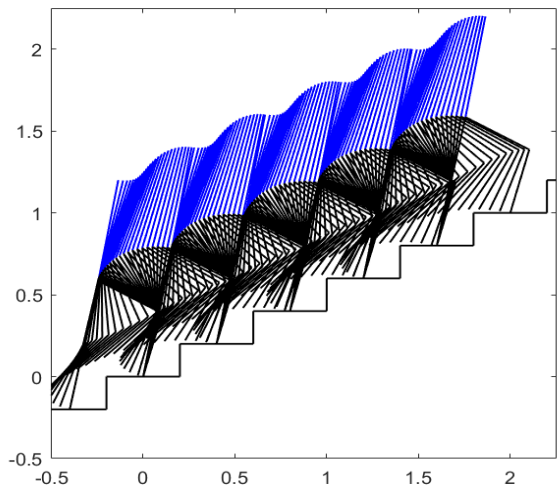
Figure 6. Stick diagram of the climbing a stair up

As can be seen in Figure 6, the robot's movement is quite normal and very similar to human movement. The trunk is kept in a good position and also the tip of the feet and other links do not touch the surfaces except at the beginning and at the end of the movement. According to the sum of the presented results, the generated path is an optimal path for the proper gait of the under-actuated bipedal robot.

**Conclusions**
In this article, we present a method to generate optimal motion for a bipedal robot, we used this method to find the paths that the 'rabbit' robot by tracking them can optimally climb stairs. This process consists of 3 parts: robot dynamic extraction (because optimization is based on the model), design of constraints based on dynamics and kinematics, and optimization. As a result of the problem, a series of virtual holonomic paths were extracted in which the zero hybrid dynamics of the problem is also satisfied, so tracking the paths are possible for under-actuated robots.

In the future, we plan to use a new method called impact invariance to design the above path, which guarantees the periodicity of the proposed paths.